\documentclass[a4paper,10pt,conference]{ieeeconf}
\IEEEoverridecommandlockouts
\usepackage{epsfig}
\usepackage{amsmath}
\usepackage{array}
\usepackage{amssymb}
\usepackage{cite}
\usepackage{booktabs}
\usepackage{multirow}
\usepackage{subcaption}
\usepackage{booktabs}
\usepackage{mathtools}
\usepackage{siunitx}
\usepackage{comment}
\usepackage{threeparttable}
\usepackage{bm}
\usepackage{pifont}
\newcommand{\cmark}{\ding{51}}%
\newcommand{\xmark}{\ding{55}}%
\makeatletter
\newcommand{\subsubsubsection}{\@startsection{paragraph}{4}{\z@}%
  {1.0\Cvs \@plus.5\Cdp \@minus.2\Cdp}%
  {.1\Cvs \@plus.3\Cdp}%
  {\reset@font\sffamily\normalsize}
}
\makeatother
\usepackage{tabularx}
\newcolumntype{Y}{>{\centering\arraybackslash}X}

\newcommand{\figlab}[1]{\label{fig:#1}}
\newcommand{\figref}[1]{Fig.~\ref{fig:#1}} 
\newcommand{\tablab}[1]{\label{tab:#1}}
\newcommand{\tabref}[1]{Table~\ref{tab:#1}} 
\newcommand{\forlab}[1]{\label{for:#1}}
\newcommand{\forref}[1]{Equation~(\ref{for:#1})} 

\usepackage{color}
\usepackage{soul}

\usepackage{amsmath}
\usepackage{algorithmicx}
\usepackage[ruled]{algorithm}
\usepackage[noend]{algpseudocode}

\begin{document}

\title{
Gait Generation Balancing Joint Load and Mobility \\for Legged Modular Robots with Easily Detachable Joints
}
\author{Kennosuke Chihara$^{1}$, Takuya Kiyokawa$^{1}$, Tetsuya Shitara$^{2}$, and Kensuke Harada$^{1,3}$
\thanks{$^{1}$Department of Systems Innovation, Graduate School of Engineering Science, The University of Osaka, 1-3 Machikaneyama, Toyonaka, Osaka, Japan.}%
\thanks{$^{2}$Graduate School of Human Sciences, The University of Osaka, 1-2 Yamadaoka, Suita, Osaka Japan.}%
\thanks{$^{3}$Industrial Cyber-Physical Systems Research Center, The National Institute of Advanced Industrial Science and Technology (AIST), 2-3-26 Aomi, Koto-ku, Tokyo, Japan.}%
}

\maketitle
\thispagestyle{empty}
\pagestyle{empty}

\begin{abstract}
Modular robots can adapt their morphology to diverse tasks, but locomotion can impose large loads on detachable joints. This paper presents an NSGA-III-based multi-objective gait optimization framework with three objectives: a locomotion speed, a static stability, and a normalized joint-load metric based on joint reaction forces. The framework yields nondominated gait candidates that reveal the trade-offs among these objectives. In a flat-terrain comparison using the 6-legged tetrapod gait, adding the joint-load objective reduced the mean maximum joint reaction force by 11.5\% and increased the mean minimum stability by 41.2\%, while decreasing the mean forward speed by 10.5\%. Simulations under flat, slope, and step conditions showed that locomotion feasibility depended on the combination of robot configuration, gait, and terrain. In physical experiments, the 4-legged trot completed locomotion on flat terrain and a $10^{\circ}$ slope, whereas the 6-legged tetrapod and tripod completed locomotion on flat terrain. These results demonstrate the feasibility of joint-load-aware gait generation based on explicit trade-offs among the three evaluated quantities.
\end{abstract}

\section{Introduction}

Modular reconfigurable robots~\cite{Gilpin2010,Yim2007,Murata2003,Zhao2020,Romiti2021,Mayer2026,Liang2026} can adapt their morphology to different tasks and environments. Representative systems include Roombots, which combines self-reconfigurable hardware with CPG-based locomotion generation~\cite{Sproewitz2009}; MoonBot, an on-demand reconfigurable platform developed for lunar infrastructure tasks~\cite{Uno2025}; and a structure-transformable humanoid equipped with an attach-lock-detachable magnetic coupling~\cite{Makabe2024}. Controller learning for directed locomotion across modular robots with different morphologies has also been studied~\cite{Lan2021}. However, gait optimization that explicitly evaluates loads transmitted through detachable interfaces has received comparatively limited attention.

In the broader field of legged locomotion, Deep Reinforcement Learning (DRL) has achieved robust locomotion over challenging terrains~\cite{Lee2020,Choi2023,Miki2024,Xu2024,Rudin2024}. Safety-oriented extensions incorporate torque constraints~\cite{Wang2025,Kim2024}, combine reinforcement learning with optimal control~\cite{Gangapurwala2022}, or use learning-based policies for robot design optimization~\cite{Belmonte2022}. Despite these advances, conflicting criteria are often combined into a scalar reward, requiring their relative weights to be specified before training. Each weighting therefore represents a particular compromise, making it difficult to explicitly examine the trade-off between joint load and locomotion performance, represented here by speed and static stability. For modular robots, the preferred compromise may also vary with robot configuration and terrain.

To address this issue, we propose an NSGA-III-based multi-objective gait optimization framework that treats the locomotion speed, static stability, and normalized joint-load metric as separate objectives. NSGA-III is adopted because its reference-point-based selection is designed to retain a diverse set of nondominated candidates across multiple objectives without combining them into a single scalar objective~\cite{Deb2013}. The overall framework is shown in \figref{3_prop_outview}. For a given robot configuration and terrain condition, the method optimizes the stride lengths, swing speeds, swing height, and duty factor. These variables were selected because stride length and swing speed govern locomotion displacement and timing, swing height governs terrain clearance and vertical leg motion, and duty factor governs the duration of body support. Together, they influence locomotion speed, static stability, and joint loading.

The main contributions of this study are threefold: (1) a three-objective gait optimization formulation that separately evaluates speed, static stability, and a normalized joint-load metric; (2) a comparison of gait optimization with and without the joint-load objective, together with an analysis of the relationship between gait parameters and the load metric; and (3) simulation and physical evaluation using 4-legged and 6-legged configurations under flat, slope, and step conditions.

\begin{figure}[tb]
    \centering
    \includegraphics[width=\linewidth]{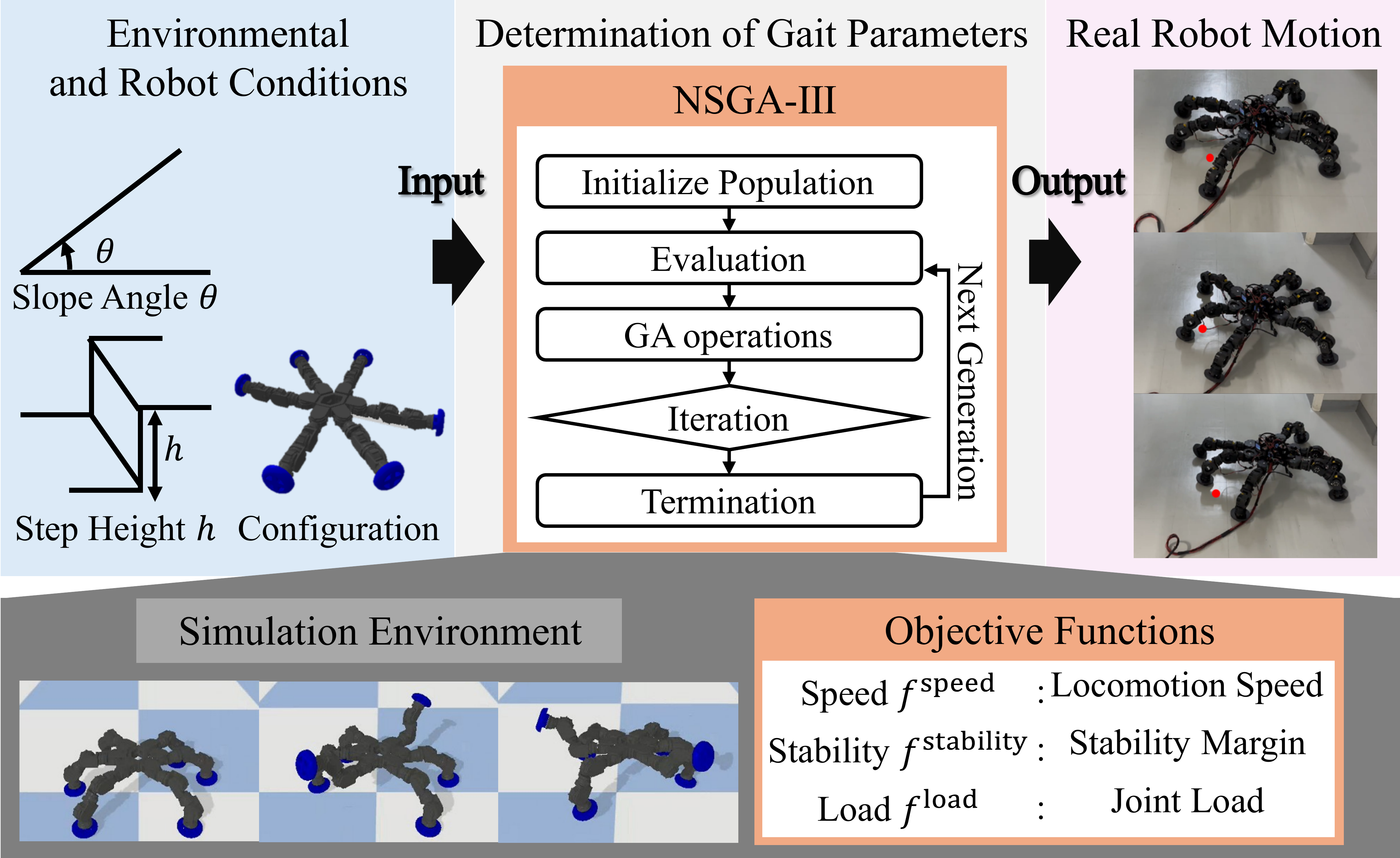}
   \caption{Overview of the proposed multi-objective gait optimization framework considering the locomotion speed, static stability, and normalized joint-load metric. The red marker in the real-robot images denotes a fixed reference point in the environment.}
    \figlab{3_prop_outview}
\end{figure}

\section{Proposed Method}

\subsection{Problem Setting}
We consider a modular robot equipped with a body for attaching module links, a control computer, rotary joint motors, and a simple detachment mechanism. A robot configuration is assumed to be selected according to the task requirements, such as the required range of motion and payload capacity, and the terrain condition. For the selected configuration, this study optimizes gait parameters using NSGA-III.

\figref{3_prop_method_robots} shows the hardware configuration of the robot used in this study, and \figref{3_prop_real_robot} illustrates the fully assembled 4-legged and 6-legged reconfigurable robots. The robot consists of modules such as rotary joints (Twister, Pivot) and end-effectors (Foot). Twister is used for leg rotation, Pivot for leg lifting, and Foot for securing grip. The joints employ a simple detachment mechanism with a physical locking structure using a coil spring and protrusion, achieving both easy detachment and high rigidity. The body frame was fabricated using a Markforged 3D printer. To achieve both light weight and high rigidity, we used Onyx, a nylon base material containing chopped carbon fibers. However, even with high-rigidity joints, there is a risk of damage in high-load tasks such as slope and step climbing. Therefore, motion generation considering joint load is essential.

\begin{figure}[tb]
    \centering
    \begin{minipage}[b]{0.37\linewidth}
        \centering
        \includegraphics[keepaspectratio, width=\linewidth]{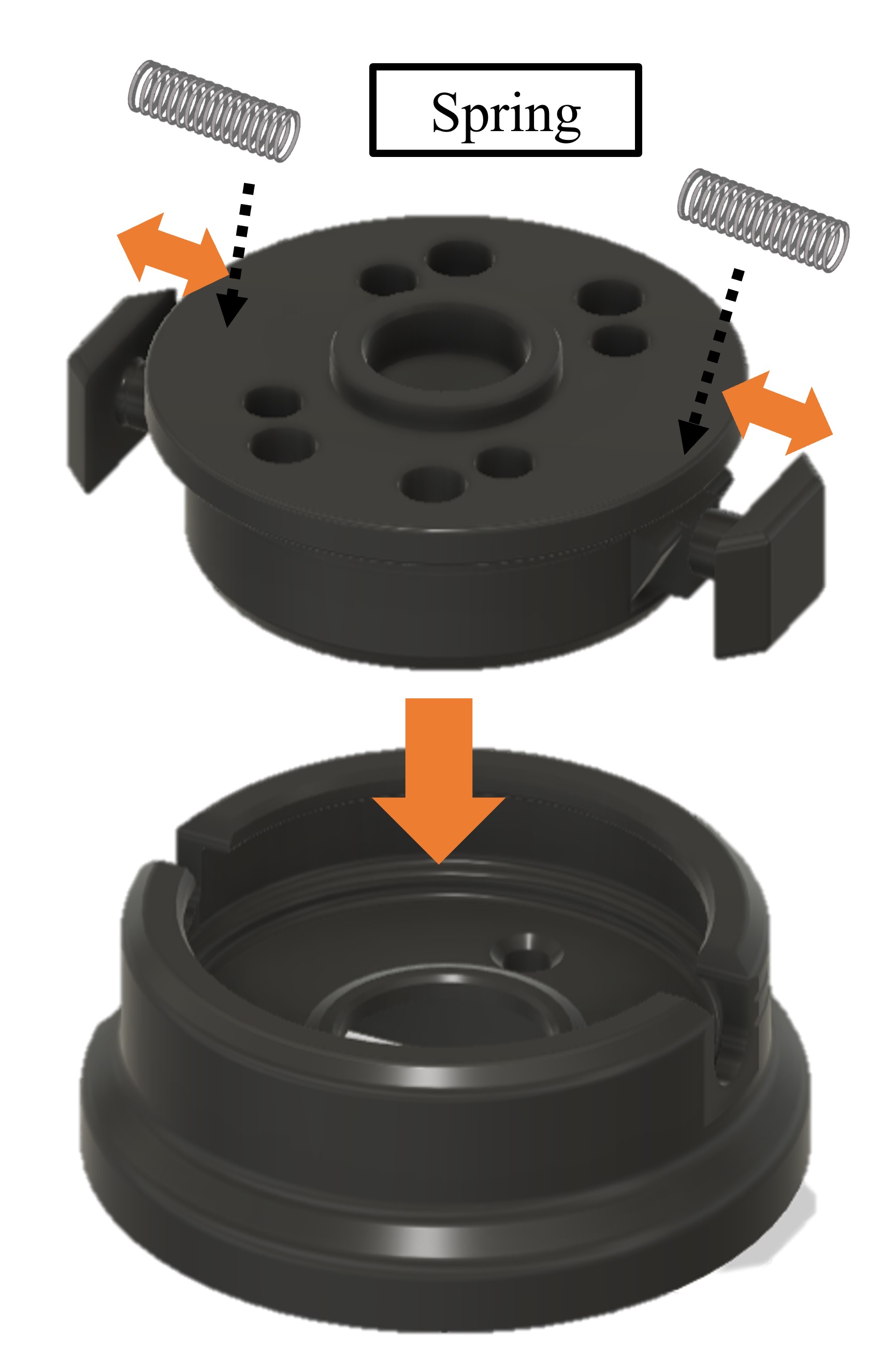}
        \subcaption{Detachable joint}
    \end{minipage}
    \begin{minipage}[b]{0.61\linewidth}
        \centering
        \includegraphics[keepaspectratio, width=\linewidth]{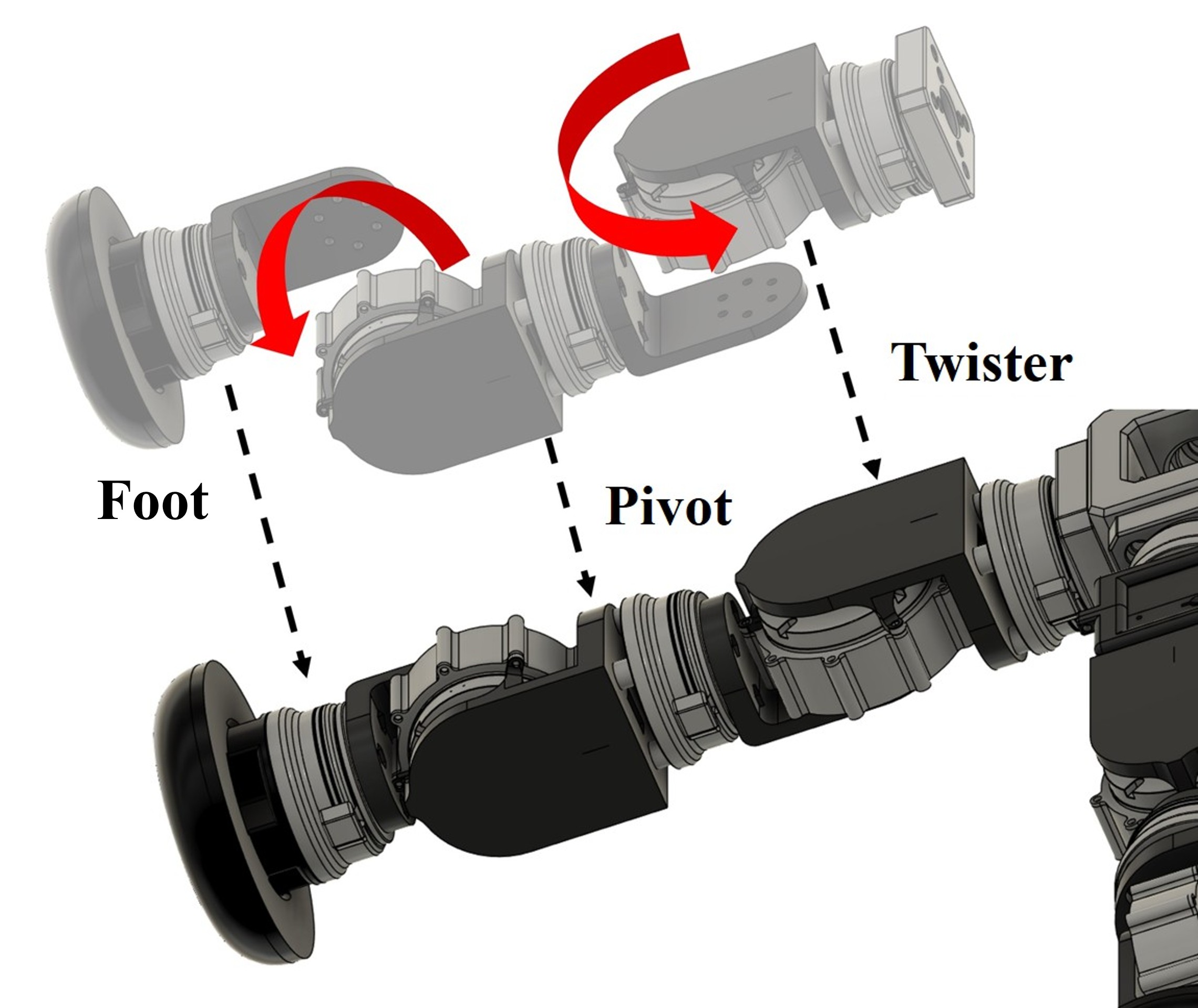}
        \subcaption{Link structure}
    \end{minipage}
    \caption{Hardware configuration including the simple detachable joint and link structure.}
    \figlab{3_prop_method_robots}
\end{figure}

\begin{figure}[tb]
    \centering
    \begin{minipage}[b]{0.49\linewidth}
        \centering
        \includegraphics[keepaspectratio, width=\linewidth]{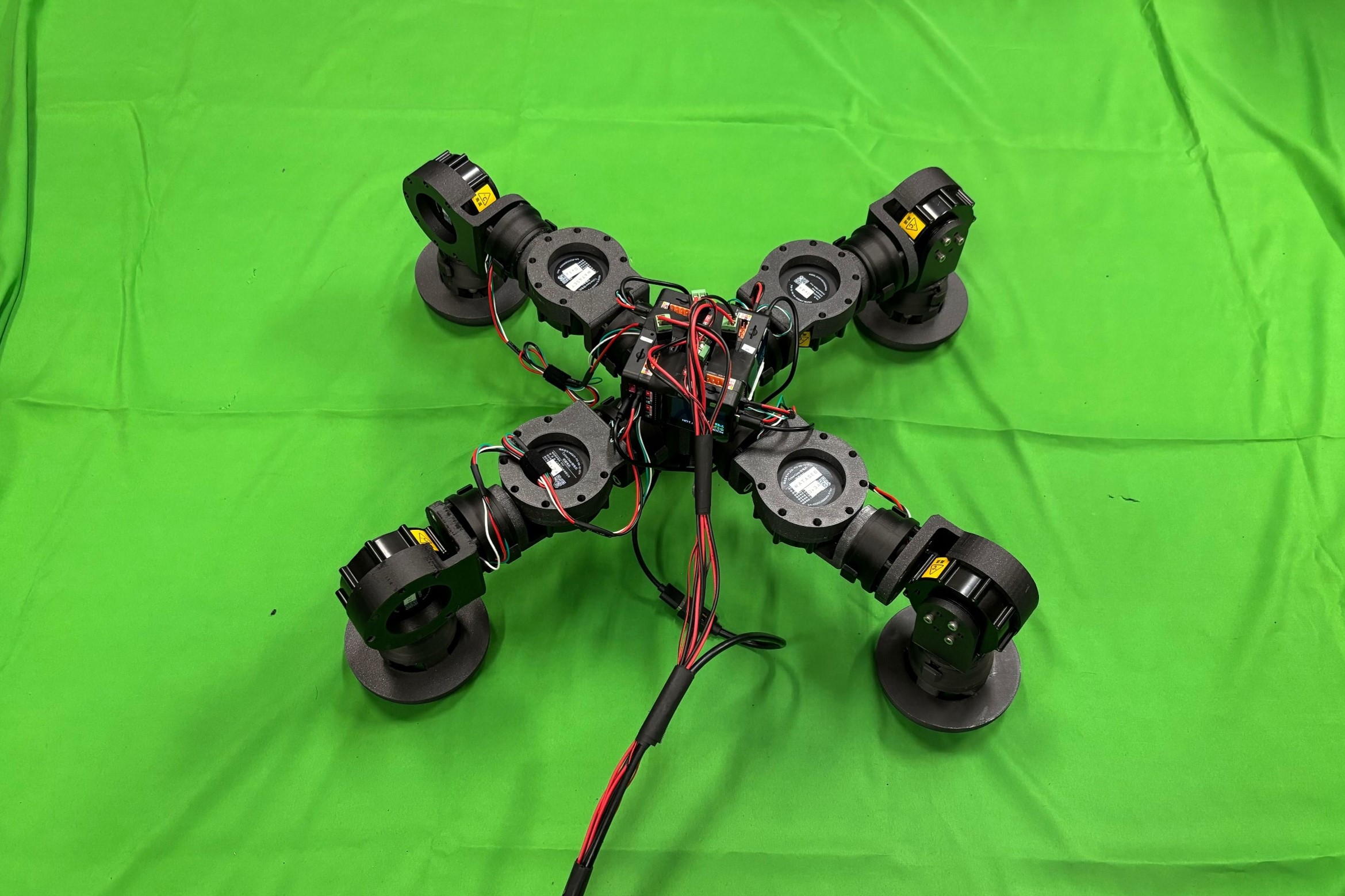}
        \subcaption{4-legged}
    \end{minipage}
    \begin{minipage}[b]{0.49\linewidth}
        \centering
        \includegraphics[keepaspectratio, width=\linewidth]{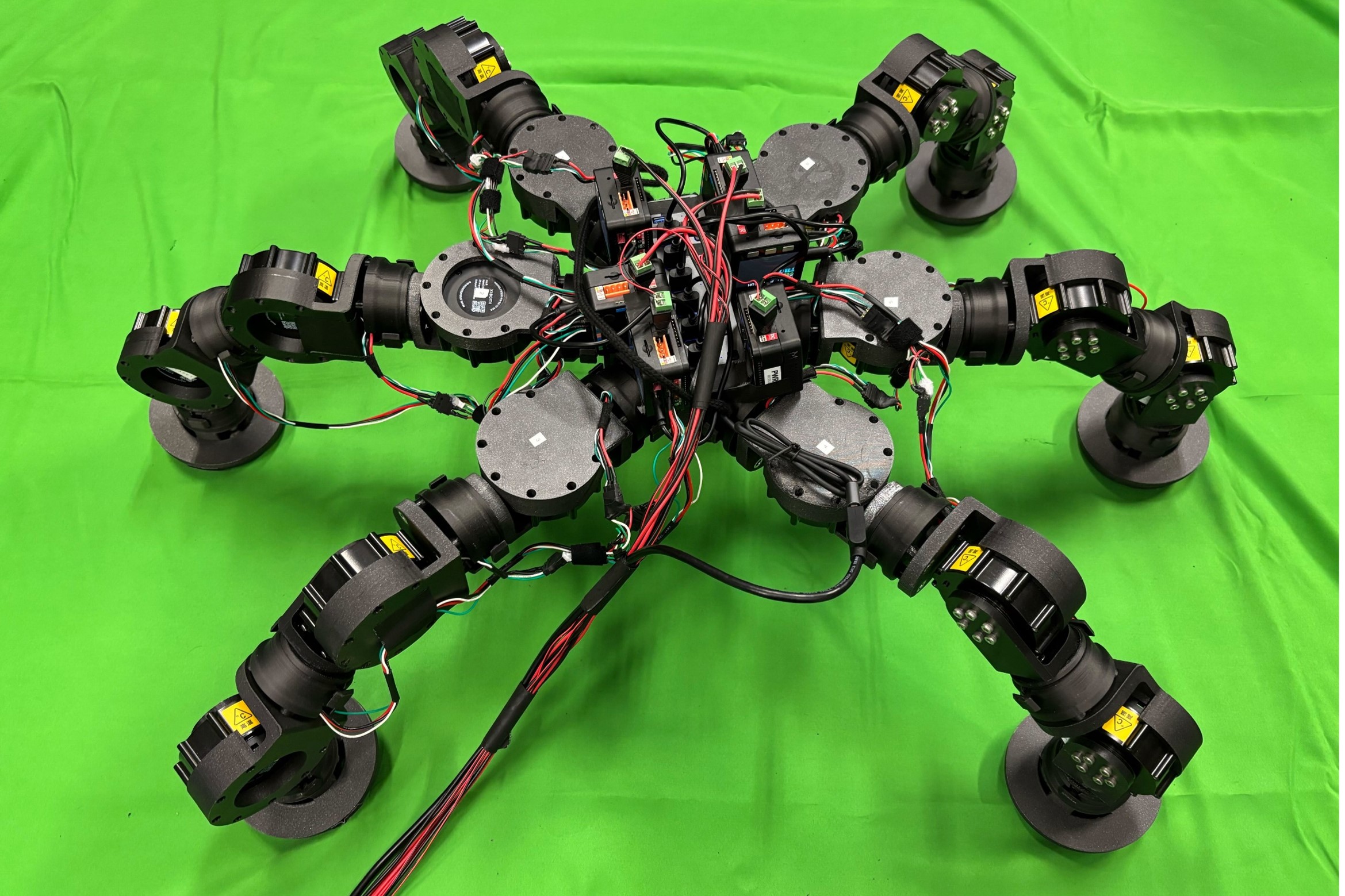}
        \subcaption{6-legged}
    \end{minipage}
    \caption{Fully assembled 4-legged and 6-legged physical robots.}
    \figlab{3_prop_real_robot}
\end{figure}

\subsection{Optimization Problem and Decision Variables}
We optimize the gait parameters using NSGA-III. The multi-objective optimization problem consists of three minimization objectives: the speed objective $f^{\text{speed}}$, stability objective $f^{\text{stability}}$, and joint-load objective $f^{\text{load}}$:

\begin{subequations}
\begin{align}
\text{minimize} \quad & \bm{J}(O) =
\begin{bmatrix}
f^{\text{speed}}(O) \\
f^{\text{stability}}(O) \\
f^{\text{load}}(O)
\end{bmatrix} \\
\text{subject to} \quad
& |\tau_j(t)| \leq \tau_\text{max},
\quad \forall j \in \mathcal{J}, \forall t
\forlab{torque_constraint} \\
& O = \{\mathbf{L}, \mathbf{V}, H, \beta\}
\end{align}
\end{subequations}

The torque constraint in~\forref{torque_constraint} restricts the exerted torque $\tau_j(t)$ within the actuator's rated limit $\tau_\text{max}$ for all active joints $\mathcal{J}$ at any time $t$. This constraint excludes candidate motions whose simulated joint torques exceed the prescribed actuator limit.

The decision variable $O$ is a set of parameters that define the leg trajectory shape, speed, and timing. Specifically, the vectors $\mathbf{L}$ and $\mathbf{V}$ denote the stride lengths and swing speeds for all legs, respectively, which include the individual stride length $L_i$ and swing speed $V_i$ for each leg $i$. Furthermore, $O$ includes the swing height $H$ common to all legs and the duty factor $\beta$. The duty factor $\beta \in (0, 1)$ represents the proportion of the stance phase within a gait cycle $T$. Since a larger $\beta$ improves stability while a smaller $\beta$ favors speed, we treat $\beta$ as a decision variable to let the optimization autonomously resolve this trade-off.

\subsection{Definition of Objective Functions}
This section describes the specific definitions of the three objective functions.
First, the speed objective $f^{\text{speed}}$ is defined as the negative normalized forward velocity with a quadratic penalty for lateral drift. Let $v_{\text{ref}}$ be the reference speed corresponding to the upper limit of the swing speed ($0.15\ \mathrm{m/s}$ in this experiment):

\begin{subequations}
    \begin{align}
        f^{\text{speed}} &= -\left(\hat{v}_{\text{fwd}} - \lambda \hat{v}_{\text{drift}}^2 \right) \\
        \hat{v}_{\text{fwd}} &= \frac{\Delta x}{T \cdot v_{\text{ref}}} \\
        \hat{v}_{\text{drift}} &= \frac{|\Delta y|}{T \cdot v_{\text{ref}}} 
    \end{align}
\end{subequations}

Here, $\hat{v}_{\text{fwd}}$ and $\hat{v}_{\text{drift}}$ are the normalized forward velocity and lateral drift velocity, respectively. $\Delta x$ and $\Delta y$ denote the forward and lateral displacements over one gait cycle $T$. Because $L_i$ and $V_i$ are optimized independently for each leg, asymmetric parameter combinations can produce lateral drift even when the forward displacement is large. The drift term therefore suppresses gait candidates that move obliquely rather than along the target direction. The quadratic form penalizes large lateral deviations more strongly than small ones. The weight $\lambda=0.5$ was selected empirically.

Next, the stability objective $f^{\text{stability}}$ is defined as the negative minimum normalized static stability margin over the evaluation period:

\begin{equation}
f^{\text{stability}} = -\min_{k \in \{1,\ldots,N_t\}} \frac{\operatorname{sdist}\!\left(\operatorname{CoM}(t_k),\operatorname{ConvexHull}(t_k)\right)}{d_{\text{nom}}}
\end{equation}

Here, $N_t$ is the number of time samples within the evaluation period, and $t_k$ denotes the $k$-th sampling time. The signed distance $\operatorname{sdist}$ is positive when the projected CoM is inside the support polygon, zero on its boundary, and negative outside it. The reference distance $d_{\text{nom}}$ is defined separately for each morphology as the inscribed-circle radius of the support polygon in the standard posture. The resulting values are $d_{\text{nom}}=0.54\ell$ for the 4-legged robot and $d_{\text{nom}}=0.62\ell$ for the 6-legged robot, where $\ell$ denotes the leg length. Thus, a larger minimum signed stability margin yields a smaller value of $f^{\text{stability}}$. If the projected CoM moves outside the support polygon at any sampled time, the minimum signed distance becomes negative and $f^{\text{stability}}$ increases.

Finally, the joint load metric $f^{\text{load}}$ is defined as the average total joint reaction force during the gait cycle, normalized by the robot's weight $W_{\text{robot}}$:

\begin{equation}
f^{\text{load}} = \frac{1} {W_{\text{robot}} N_t |\mathcal{J}|} \sum_{k=1}^{N_t} \sum_{j \in \mathcal{J}} \left\| \bm{F}_j(t_k) \right\|_2
\end{equation}

where $\bm{F}_j(t_k)$ is the reaction-force vector acting on joint $j$ at time $t_k$, and $|\mathcal{J}|$ is the number of evaluated joints. This metric is the mean joint-reaction-force magnitude over all sampled joint--time pairs, normalized by the robot weight $W_{\text{robot}}$. Minimizing it favors gait candidates with lower evaluated interface loads.

For reporting and visualization, the normalized speed is defined as $-f^{\text{speed}}$, whereas the minimum stability is defined as $-f^{\text{stability}}$. A negative minimum stability therefore indicates that the projected CoM moved outside the support polygon at least once during the evaluation period.

\section{Experiments}
\subsection{Overview}
The proposed framework is evaluated through simulations and physical experiments in four parts: (1) a comparison of formulations with and without the joint-load objective (Section~III-B); (2) an evaluation of gait feasibility across robot configurations and terrains (Section~III-C); (3) a multiple-regression analysis of the relationship between gait parameters and the joint-load metric (Section~III-D); and (4) a physical-robot evaluation of selected gait candidates (Section~III-E).

We constructed 4-legged and 6-legged robot models using the physics engine PyBullet, with mass properties matched to those of the corresponding physical robots. To evaluate the performance of these models, we prepared three types of environmental conditions illustrated in \figref{4_exp_environment}: flat terrain, a slope with a $10^{\circ}$ inclination, and a step with a height of $10\ \mathrm{cm}$. In these evaluations, a trial is considered a failure if the robot falls on the flat or slope terrains, or if it fails to bring its entire body onto the upper level in the step environment. Conversely, any trial that fulfills the locomotion task without satisfying these failure criteria is defined as a success.

\begin{figure}[tb]
    \centering
    \begin{minipage}[b]{0.32\linewidth}
        \centering
        \includegraphics[keepaspectratio, width=\linewidth]{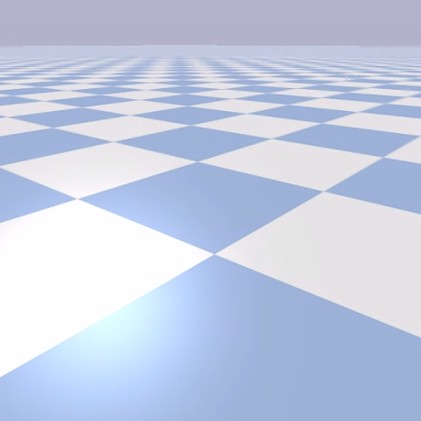}
        \subcaption{Flat}
    \end{minipage}
    \begin{minipage}[b]{0.32\linewidth}
        \centering
        \includegraphics[keepaspectratio, width=\linewidth]{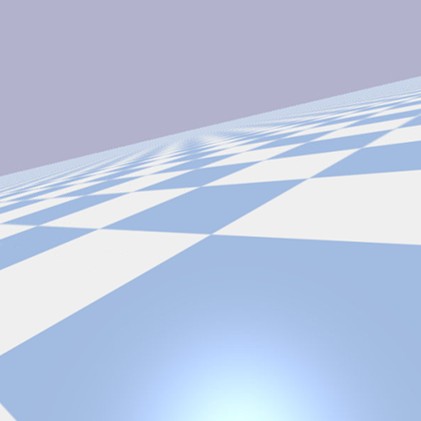}
        \subcaption{Slope}
    \end{minipage}
    \begin{minipage}[b]{0.32\linewidth}
        \centering
        \includegraphics[keepaspectratio, width=\linewidth]{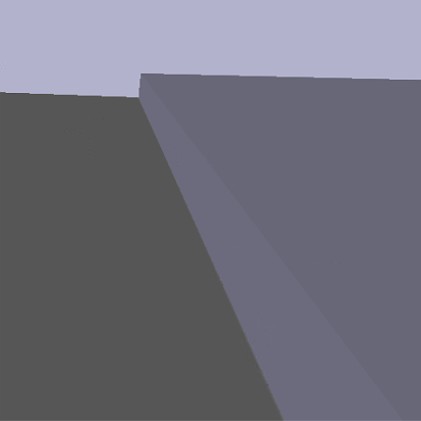}
        \subcaption{Step}
    \end{minipage}
    \caption{Simulated environmental conditions: flat terrain, $10^{\circ}$ slope, and $10\ \mathrm{cm}$ step.}
    \figlab{4_exp_environment}
\end{figure}

\begin{table}[tb]
  \centering
  \caption{NSGA-III parameter settings.}
  \tablab{4_exp_nsga3_params}
  \renewcommand{\arraystretch}{1.1}
  \begin{tabular}{lc} \toprule
    Parameter & Value \\ \midrule
    Objectives ($M$) & 3 \\
    Population size ($N_{\mathrm{pop}}$) & 91 \\
    Generations ($G$) & 10 \\
    Decision variables ($n_{\mathrm{var}}$) & $2N_{\mathrm{leg}}+2$ \\
    Crossover ($P_c, \eta_c$) & 1.0, 30 \\
    Mutation ($P_m, \eta_m$) & $1/n_{\mathrm{var}}$, 20 \\ \bottomrule
  \end{tabular}
\end{table}

\begin{table}[tb]
  \centering
  \caption{Search range for decision variables.}
  \tablab{4_exp_decision_bounds}
  \begin{tabular}{lllc} \toprule
    Variable & Unit & Gait & Range \\ \midrule
    Stride length $L_i$ & $\mathrm{m}$ & All & $[0.05, 0.30]$ \\
    Swing speed $V_i$ & $\mathrm{m/s}$ & All & $[0.01, 0.15]$ \\
    Swing height $H$ & $\mathrm{m}$ & All & $[0.10, 0.50]$ \\
    Duty factor $\beta$ & - & Wave & $[0.84, 0.95]$ \\
     & & Tetrapod & $[0.67, 0.85]$ \\
     & & Trot & $[0.51, 0.70]$ \\
     & & Tripod & $[0.51, 0.70]$ \\ \bottomrule
  \end{tabular}
\end{table}

The NSGA-III settings are summarized in~\tabref{4_exp_nsga3_params}. The number of objectives was set to $M=3$, corresponding to the speed, minimum-stability, and joint-load objectives. The population size $N_{\mathrm{pop}}$ was set to 91 according to the number of reference points, and the optimization was performed for $G=10$ generations. Let $N_{\mathrm{leg}}$ denote the number of legs. Because each leg has an individual stride length $L_i$ and swing speed $V_i$, together with the common swing height $H$ and duty factor $\beta$, the number of decision variables was $n_{\mathrm{var}}=2N_{\mathrm{leg}}+2$. This resulted in 10 variables for the 4-legged robot and 14 variables for the 6-legged robot. The crossover probability was set to $P_c=1.0$, meaning that crossover was applied to every selected parent pair, and the crossover distribution index was set to $\eta_c=30$. The mutation probability was set to $P_m=1/n_{\mathrm{var}}$ for each decision variable, corresponding to one mutated variable on average per offspring, and the mutation distribution index was set to $\eta_m=20$. The distribution indices $\eta_c$ and $\eta_m$ control the spread of the offspring generated by crossover and mutation, respectively.

The gait pattern was specified by assigning a normalized phase offset $\phi_i \in [0,1)$ to the periodic trajectory of each leg $i$. A phase difference of $0.5$ between two legs corresponds to a half-cycle shift. The phase offset and duty factor have different roles: $\phi_i$ determines the relative timing among the legs, whereas the duty factor $\beta$ determines the fraction of each leg's cycle spent in the stance phase. The phase offsets were fixed for each gait, whereas $\beta$ was optimized.

For the 4-legged configuration, only the trot gait was evaluated. The diagonal leg pairs $\{1,4\}$ and $\{2,3\}$ were assigned phase offsets of $0$ and $0.5$, respectively. For the 6-legged configuration, the wave, tetrapod, and tripod gaits were evaluated. In the wave gait, the phase offsets increase successively by $1/6$, causing the legs to enter the swing phase one at a time. In the tetrapod gait, three leg pairs are assigned phase offsets of $0$, $1/3$, and $2/3$. In the tripod gait, the two leg groups $\{1,4,5\}$ and $\{2,3,6\}$ are assigned phase offsets of $0$ and $0.5$, respectively, producing alternating three-leg support.

The search ranges of the decision variables are summarized in~\tabref{4_exp_decision_bounds}. The ranges of $L_i$, $V_i$, and $H$ were selected empirically to cover executable gait motions while excluding nearly zero or excessively large motions. The stride-length range $L_i\in[0.05,0.30]\ \mathrm{m}$ covers short to long strides, while the swing-speed range $V_i\in[0.01,0.15]\ \mathrm{m/s}$ covers slow to fast leg motions; its upper bound is also used as the reference speed $v_{\mathrm{ref}}$. The swing-height range $H\in[0.10,0.50]\ \mathrm{m}$ includes the clearance required for the $0.10\ \mathrm{m}$ step and allows the optimization to explore larger swing motions.

The lower bounds of the duty-factor ranges were determined from the fixed phase pattern of each gait. For the wave gait, the lower bound $\beta=0.84$ is slightly larger than $5/6$, limiting the swing phase to at most one leg at a time and leaving at least five supporting legs. For the tetrapod gait, the lower bound $\beta=0.67$ is slightly larger than $2/3$, limiting the swing phase to one leg pair at a time and leaving at least four supporting legs. For the trot and tripod gaits, the lower bound $\beta=0.51$ is slightly larger than $0.5$, creating a short overlap between the two alternating support groups. The upper bounds were selected empirically to retain sufficient swing duration while allowing the optimizer to vary the balance between support stability and locomotion speed.

Additionally, an absolute torque limit of $12\ \mathrm{Nm}$ was applied to all actuators (Xiaomi CyberGear). This threshold was determined not only by the motor's rated capacity but also to prevent the physical locking structure of the detachable joints from disengaging under excessive stress. The simulation in PyBullet was performed with a control frequency of 240 Hz.

For physical verification, we constructed 4-legged and 6-legged robots with kinematic configurations equivalent to those used in the simulation. The trajectory generated by an external PC was sent to the M5Stack Core2 on each leg via USB serial communication, and the actuators were controlled via CAN communication. Each leg was powered independently to prevent interference caused by voltage drops.
For visual comparison across frames within each motion sequence, the red marker denotes a fixed reference point in the environment; its apparent image position may vary with the camera viewpoint.

\subsection{Performance Changes due to Joint Load Consideration}
First, the two-objective formulation without the joint-load objective and the proposed three-objective formulation were compared for the 6-legged tetrapod gait on flat terrain. Both formulations used the same NSGA-III settings. The two-objective formulation minimized only $f^{\text{speed}}$ and $f^{\text{stability}}$, whereas the proposed formulation additionally minimized $f^{\text{load}}$. This comparison isolates the effect of adding the joint-load objective. To evaluate the overall tendency rather than individual solutions, the objective values were averaged over the final-generation nondominated set for each formulation. For the load, the maximum joint reaction force of each solution, $\max_{k,j}\|\bm{F}_j(t_k)\|_2$, was evaluated and averaged over the set. The mean maximum joint reaction force decreased from $645\ \mathrm{N}$ to $571\ \mathrm{N}$ (approximately 11.5\%), and the mean of the minimum-stability values increased by approximately 41.2\%.

As shown in \figref{5_exp_sim1_merged}, the gait obtained with the joint-load objective has a smaller swing height than that obtained without it. The mean forward speed decreased by approximately 10.5\%.

\begin{table}[tb]
 \centering
 \caption{Feasibility of locomotion in simulation experiments.}
 \tablab{5_exp_result_matrix}
 \begin{tabularx}{\linewidth}{llYYY} \toprule
  Legs & Gait & Flat & Slope & Step \\ \midrule
  4 & Trot & \cmark & \cmark & - \\ \midrule
  6 & Wave & \xmark & \xmark & \xmark \\
    & Tetrapod & \cmark & \cmark & \xmark \\
    & Tripod & \cmark & \xmark & \cmark \\ \bottomrule
  \multicolumn{5}{l}{\footnotesize \cmark: Feasible gait found, \xmark: None found, -: Not executed} \\
 \end{tabularx}
\end{table}

\begin{figure}[tb]
  \centering
  \begin{minipage}[b]{\linewidth}
    \centering
    \includegraphics[width=\linewidth]{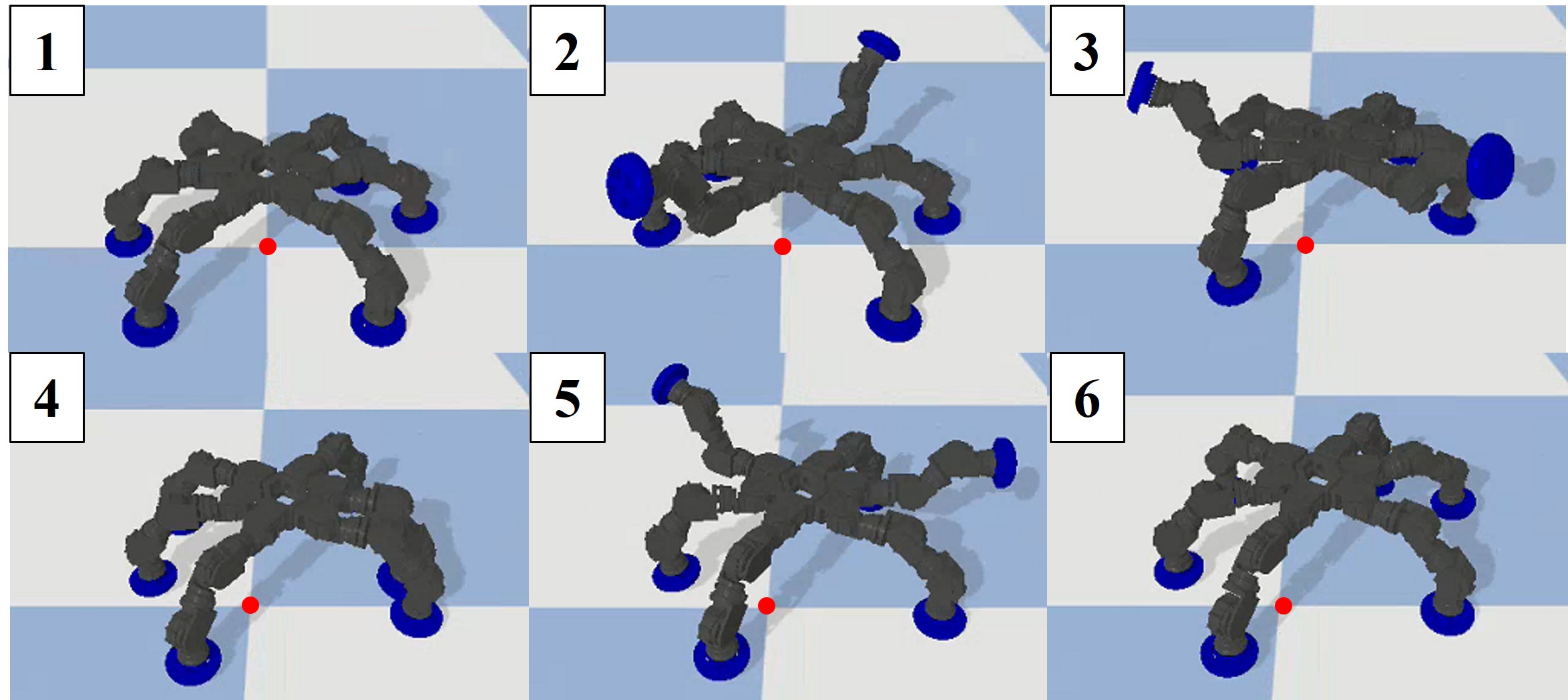}
    \subcaption{Without the joint-load objective}
    \figlab{5_exp_sim1_without_load}
  \end{minipage}
  \begin{minipage}[b]{\linewidth}
    \centering
    \includegraphics[width=\linewidth]{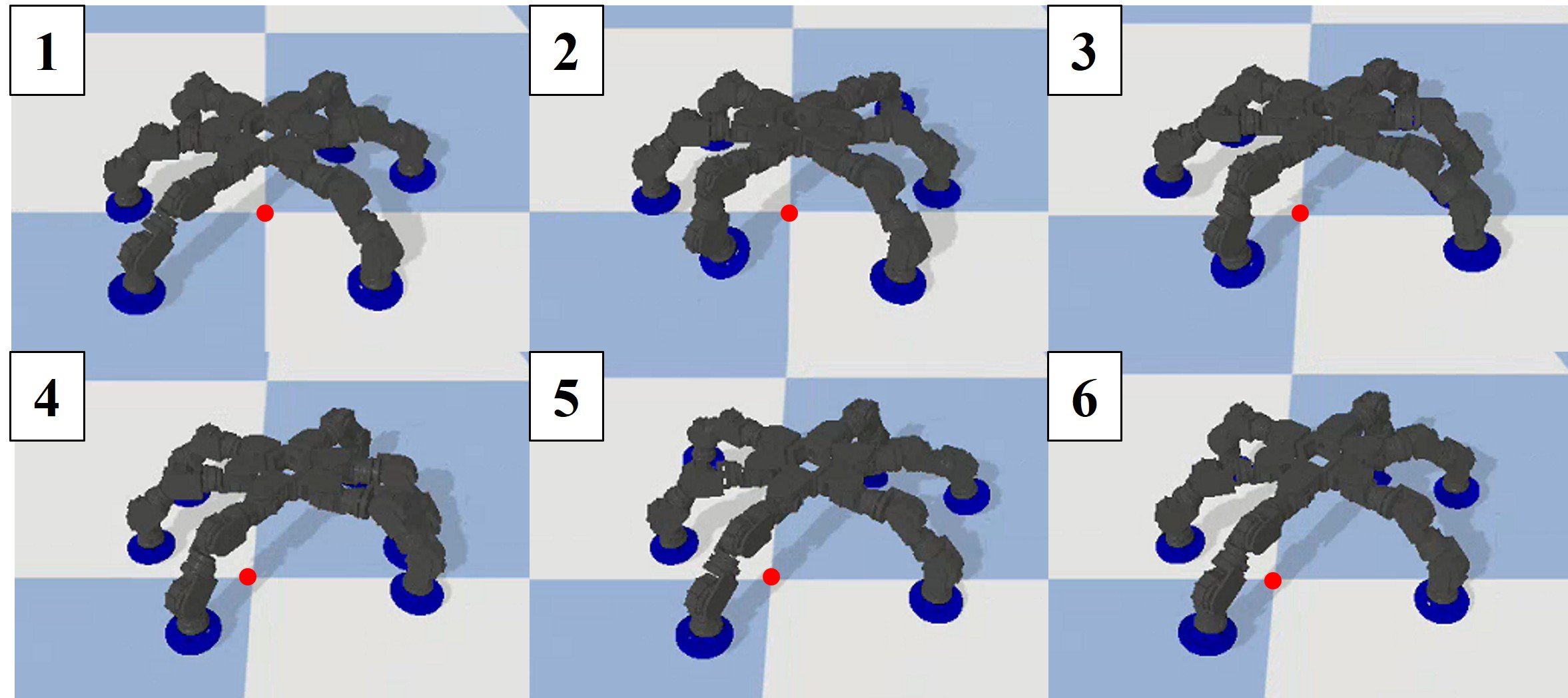}
    \subcaption{With the joint-load objective}
    \figlab{5_exp_sim1_with_load}
  \end{minipage}
  \caption{Comparison of locomotion motions with and without joint load consideration for the 6-legged tetrapod gait on flat terrain.}
  \figlab{5_exp_sim1_merged}
\end{figure}

\subsection{Adaptive Capabilities Analysis by Environment and Gait}
Next, we verified the adaptability of each gait for the 4-legged and 6-legged robots in flat, slope, and step environments. \tabref{5_exp_result_matrix} indicates whether at least one feasible gait was found for each condition. A condition was marked as successful when an obtained gait completed the locomotion task; otherwise, it was marked as failed. Thus, the table reports feasibility rather than a statistical success rate over repeated independent runs.

On flat terrain, feasible locomotion was obtained for all gaits except the 6-legged wave gait. Among the successful gaits, the 6-legged tetrapod gait achieved the highest normalized speed of 0.508 (forward speed: $0.114\ \mathrm{m/s}$). A representative high-stability solution for the 4-legged trot gait had a normalized speed of 0.203 (forward speed: $0.046\ \mathrm{m/s}$) and a minimum stability of 0.509; the large duty factor selected by the optimization increased the duration of four-leg support. The 6-legged wave gait exhibited severe body oscillation and a negative minimum stability of $-0.252$.

\begin{figure*}[tb]
  \centering
  \begin{minipage}[tb]{0.49\textwidth}
    \centering
    \includegraphics[width=\linewidth]{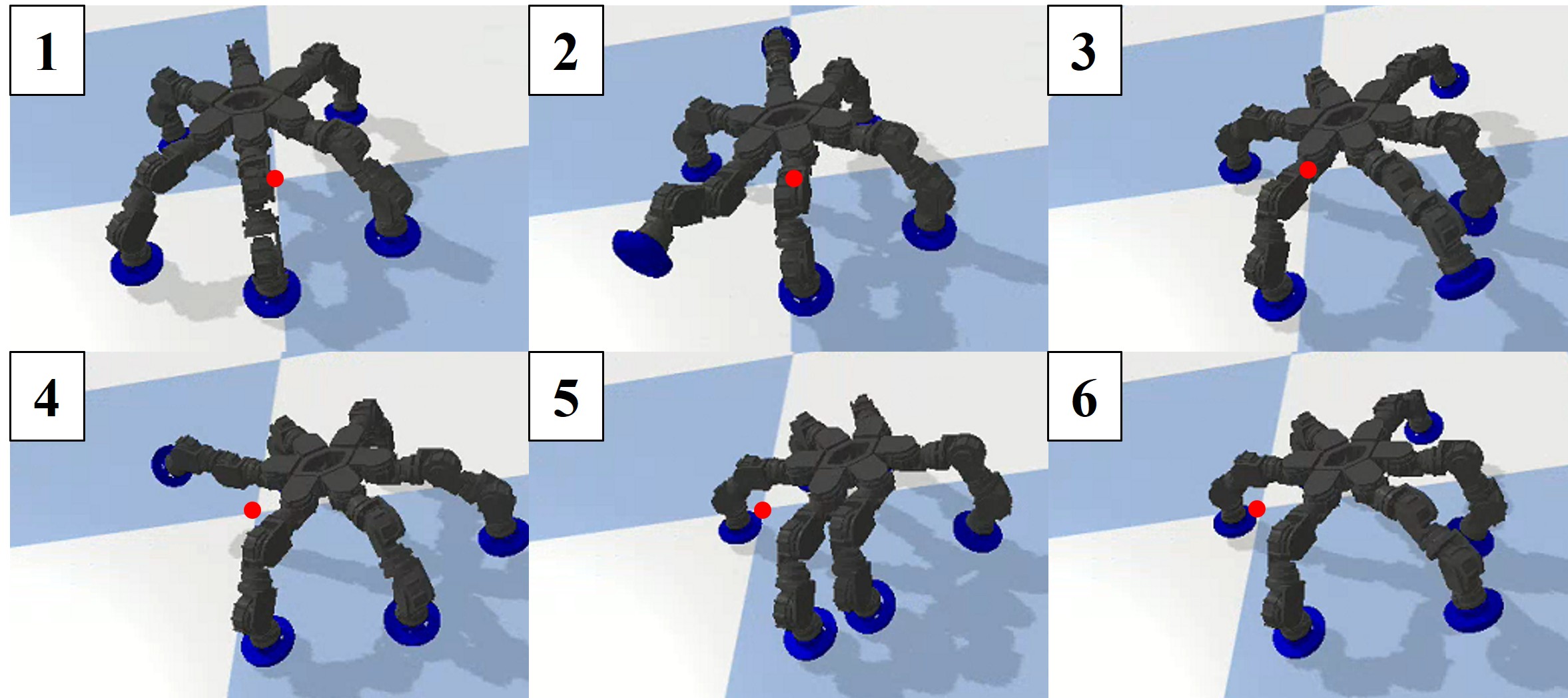}
    \subcaption{6-legged tetrapod gait on the slope}
    \figlab{5_exp_tetrapod_slope_sequence}
  \end{minipage}
  \hfill
  \begin{minipage}[tb]{0.49\textwidth}
    \centering
    \includegraphics[width=\linewidth]{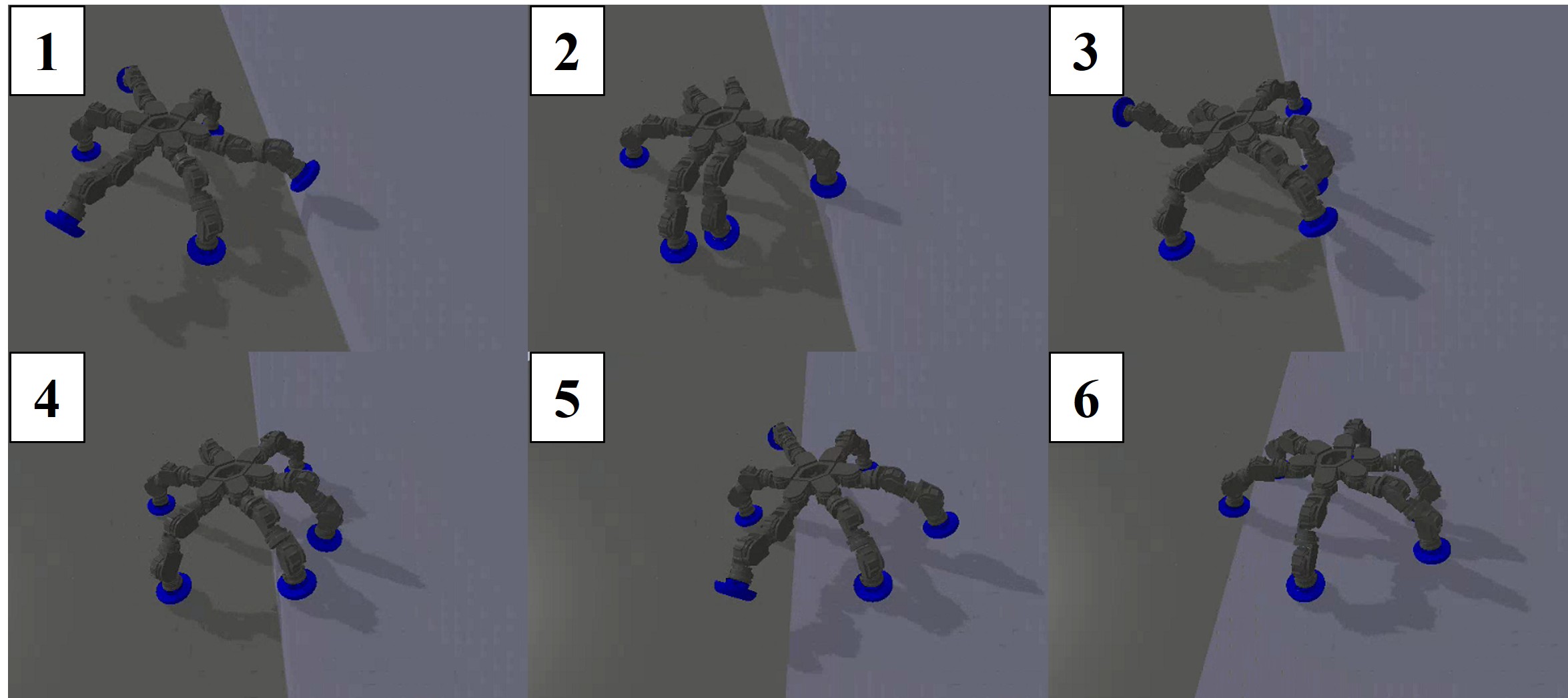}
    \subcaption{6-legged tripod gait on the step}
    \figlab{5_exp_tripod_step_sequence}
  \end{minipage}
  \caption{Representative successful locomotion sequences obtained in simulation for uneven environments: (a) the 6-legged tetrapod gait on the $10^{\circ}$ slope and (b) the 6-legged tripod gait over the $10\ \mathrm{cm}$ step. The red marker denotes a fixed reference point in the environment.}
  \figlab{5_exp_uneven_gait_sequences}
\end{figure*}

\figref{5_exp_uneven_gait_sequences} shows representative successful locomotion sequences obtained for the 6-legged robot in the uneven environments. On the slope, the tetrapod gait maintains at least four supporting legs, forming a relatively large support polygon and favoring static stability. In the step environment, the tripod gait alternates between two groups of three supporting legs. Although this results in a smaller support polygon, simultaneously swinging the other three legs allows them to be lifted and repositioned to clear the step. These results indicate that the appropriate balance between support stability and swing-leg mobility depends on the terrain.

\begin{figure}[tb]
  \centering
  \begin{minipage}[b]{0.48\linewidth}
    \centering
    \includegraphics[width=\linewidth]{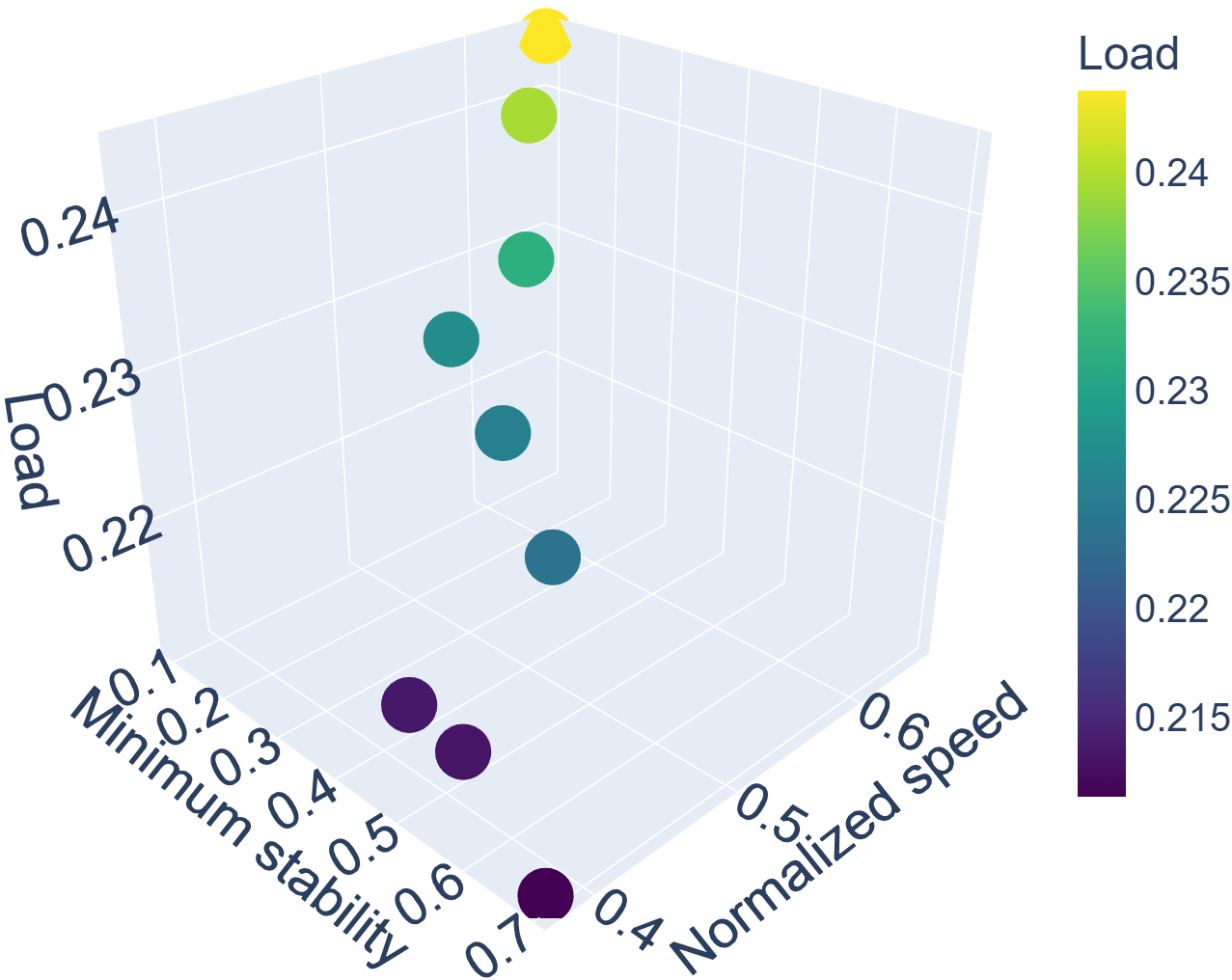}
    \subcaption{6-legged Tetrapod (Success)}
    \figlab{5_exp_pareto_tetrapod}
  \end{minipage}
  \hfill
  \begin{minipage}[b]{0.48\linewidth}
    \centering
    \includegraphics[width=\linewidth]{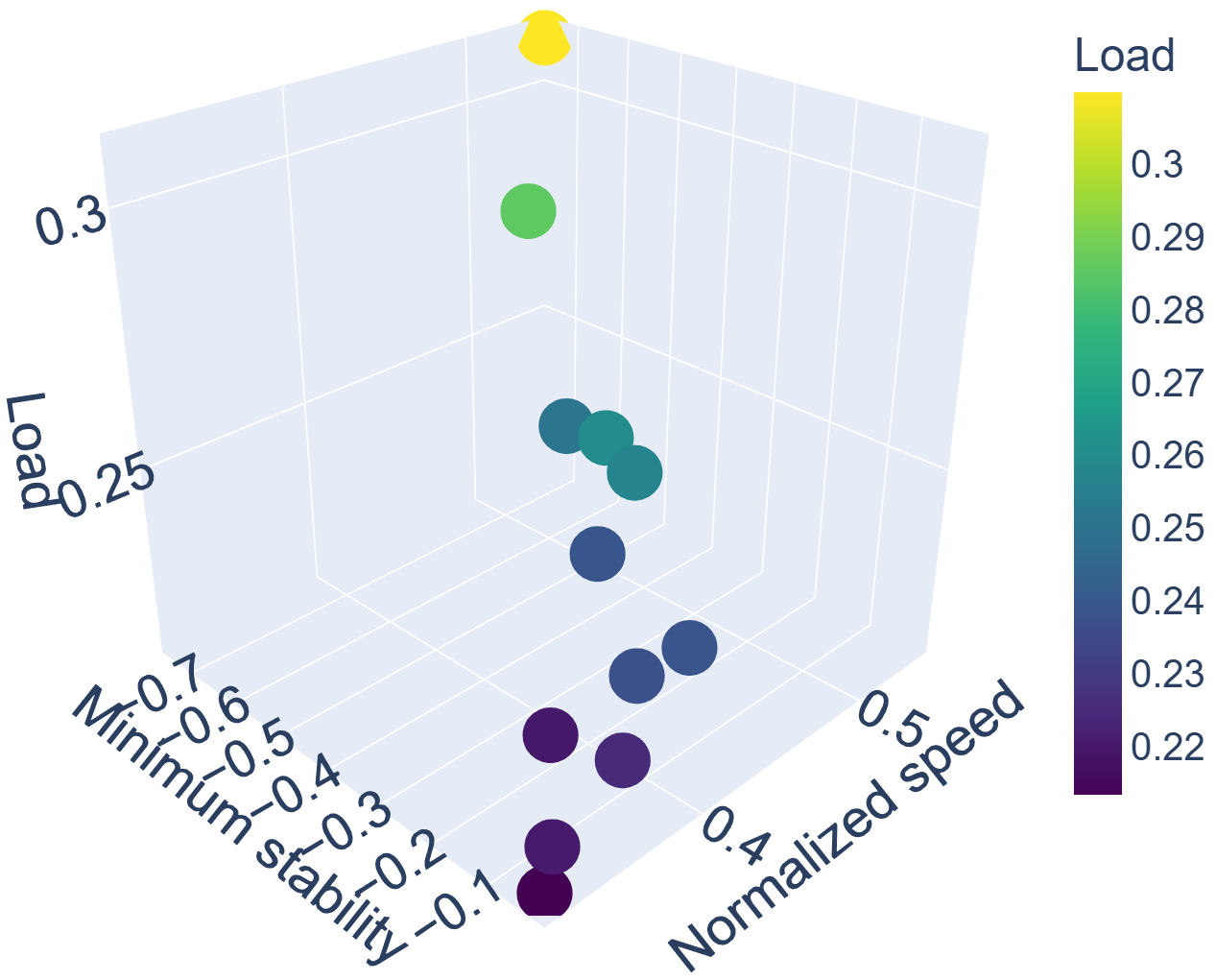}
    \subcaption{6-legged Wave (Failure)}
    \figlab{5_exp_pareto_wave}
  \end{minipage}
  \caption{Comparison of the obtained nondominated solutions on flat terrain. Point color also represents the joint-load metric shown on the vertical axis. (a) The tetrapod gait achieves successful locomotion with solutions distributed in the positive-minimum-stability region. (b) The wave gait fails as most solutions are concentrated in the negative-minimum-stability region.}
  \figlab{5_exp_pareto_fronts}
\end{figure}

To examine the relationship between the objective values and locomotion feasibility, \figref{5_exp_pareto_fronts} plots the nondominated solutions for the tetrapod and wave gaits in the three-dimensional objective space defined by the locomotion speed, minimum stability, and joint-load metric. For the tetrapod gait, solutions with positive minimum stability values span a range of speed and joint-load values, allowing selection according to the desired trade-off. In contrast, most wave-gait solutions have negative minimum stability values, consistent with the observed locomotion failure.

\subsection{Statistical Analysis of Factors Affecting Joint Load}
To identify the specific factors influencing joint load, we conducted a multiple regression analysis using the acquired nondominated solutions. The explanatory variables were classified into three groups: (1) duty factor $\beta$ and swing height $H$, (2) stride length of each leg ($L_{1}$ to $L_{N_\mathrm{leg}}$), and (3) swing speed of each leg ($V_{1}$ to $V_{N_\mathrm{leg}}$).

The analysis on flat terrain showed statistically significant positive regression coefficients for the duty factor and swing height ($p < 0.05$). For swing height, the regression coefficient was $B=0.401$ ($p=0.004$). Conversely, none of the variables related to the stride length and swing speed of individual legs had a statistically significant regression coefficient ($p > 0.05$).

When the same multiple regression analysis was applied to the nondominated solutions obtained in the slope and step environments, none of the explanatory variables had a statistically significant regression coefficient ($p > 0.05$).

\begin{table}[tb]
 \centering
 \caption{Observed feasibility in physical locomotion experiments.}
 \tablab{5_exp_real_result_matrix}
 \begin{tabularx}{\linewidth}{llYYY} \toprule
  Legs & Gait & Flat & Slope & Step \\ \midrule
  4 & Trot & \cmark & \cmark & - \\ \midrule
  6 & Wave & - & - & - \\
    & Tetrapod & \cmark & \xmark & - \\
    & Tripod & \cmark & \xmark & \cmark$^{*}$ \\ \bottomrule
  \multicolumn{5}{l}{\footnotesize \cmark: Success, \xmark: Failure, \cmark$^{*}$: Conditional success with support} \\
 \end{tabularx}
\end{table}

\begin{figure}[tb!]
  \centering
  \begin{minipage}[tb]{\linewidth}
    \centering
    \includegraphics[width=\linewidth]{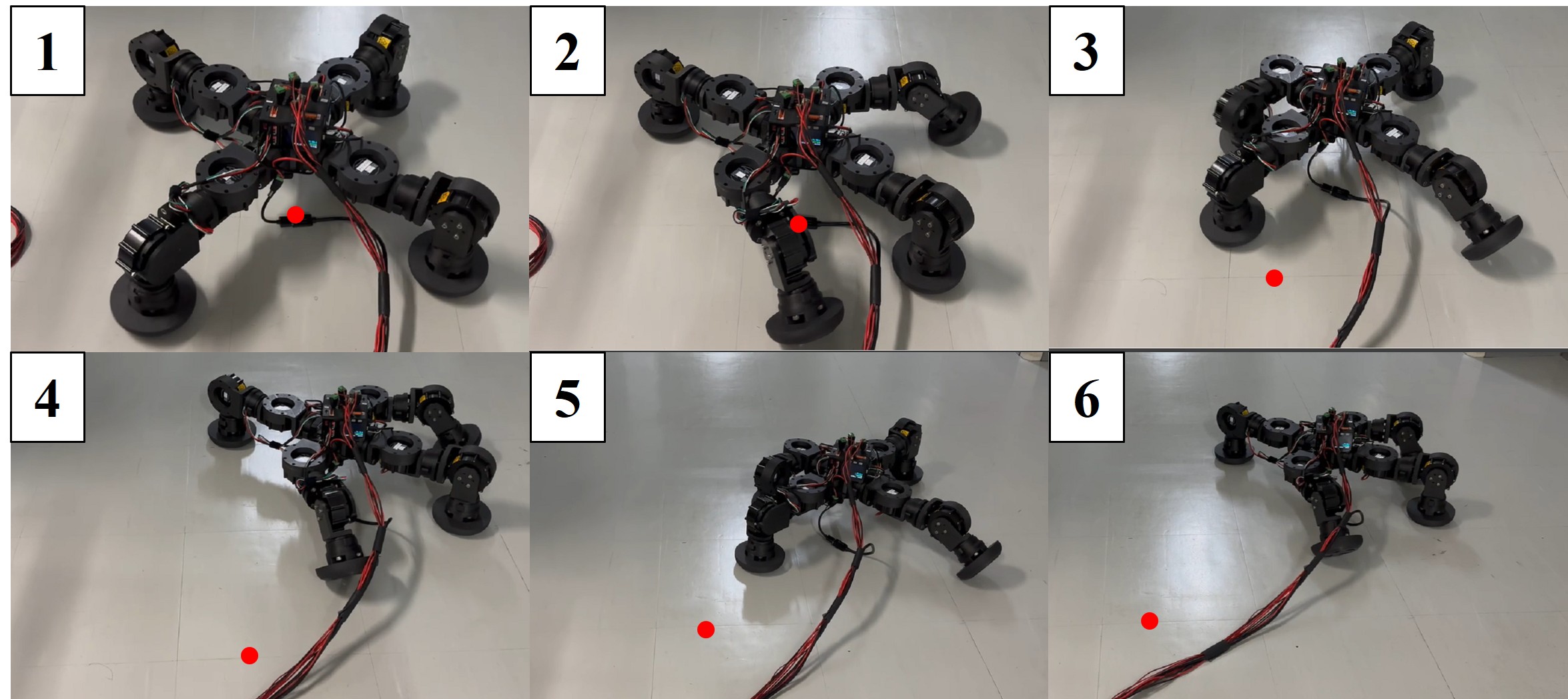}
    \subcaption{4-legged trot (flat)}
  \end{minipage}
  \hfill
  \begin{minipage}[tb]{\linewidth}
    \centering
    \includegraphics[width=\linewidth]{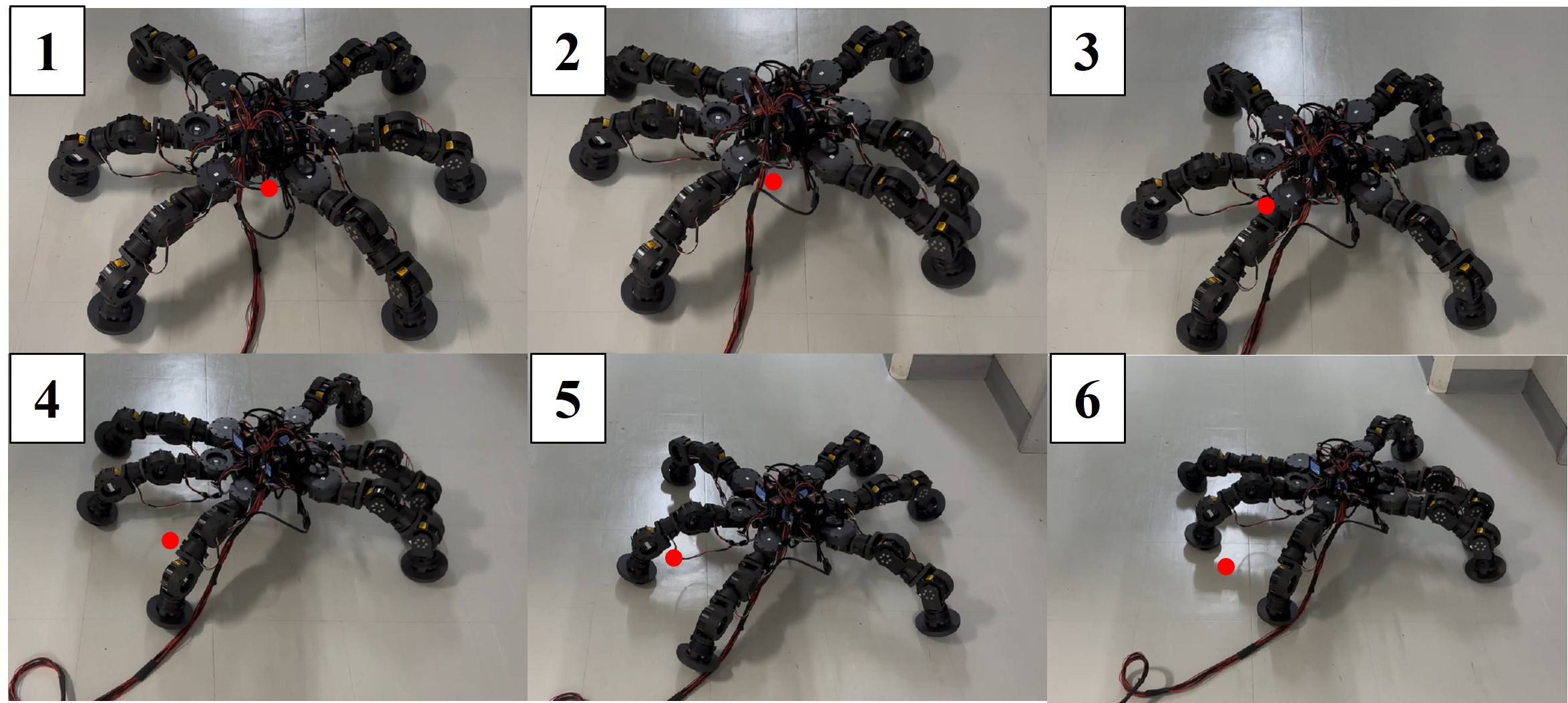}
    \subcaption{6-legged tetrapod (flat)}
  \end{minipage}
  \hfill
  \begin{minipage}[tb]{\linewidth}
    \centering
    \includegraphics[width=\linewidth]{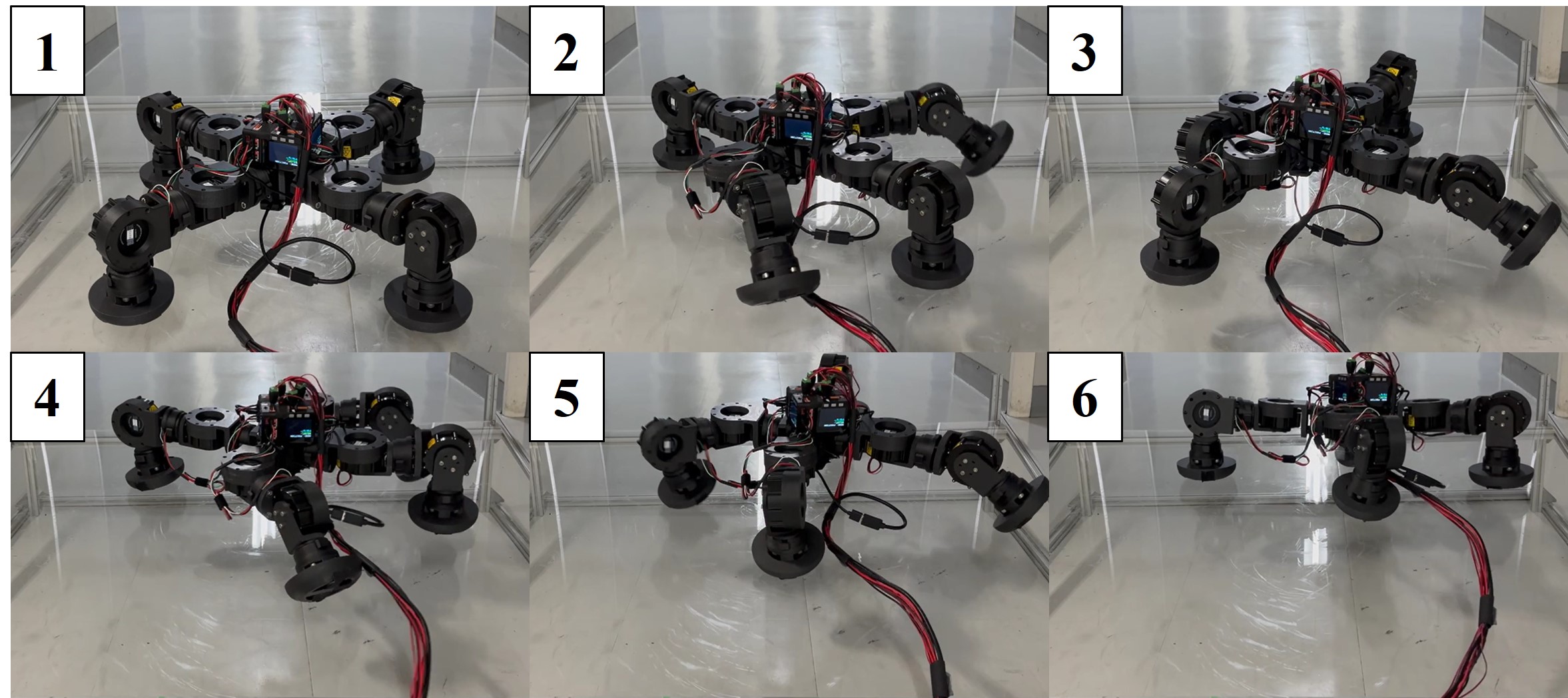}
    \subcaption{4-legged trot (slope)}
  \end{minipage}
  \hfill
  \begin{minipage}[tb]{\linewidth}
    \centering
    \includegraphics[width=\linewidth]{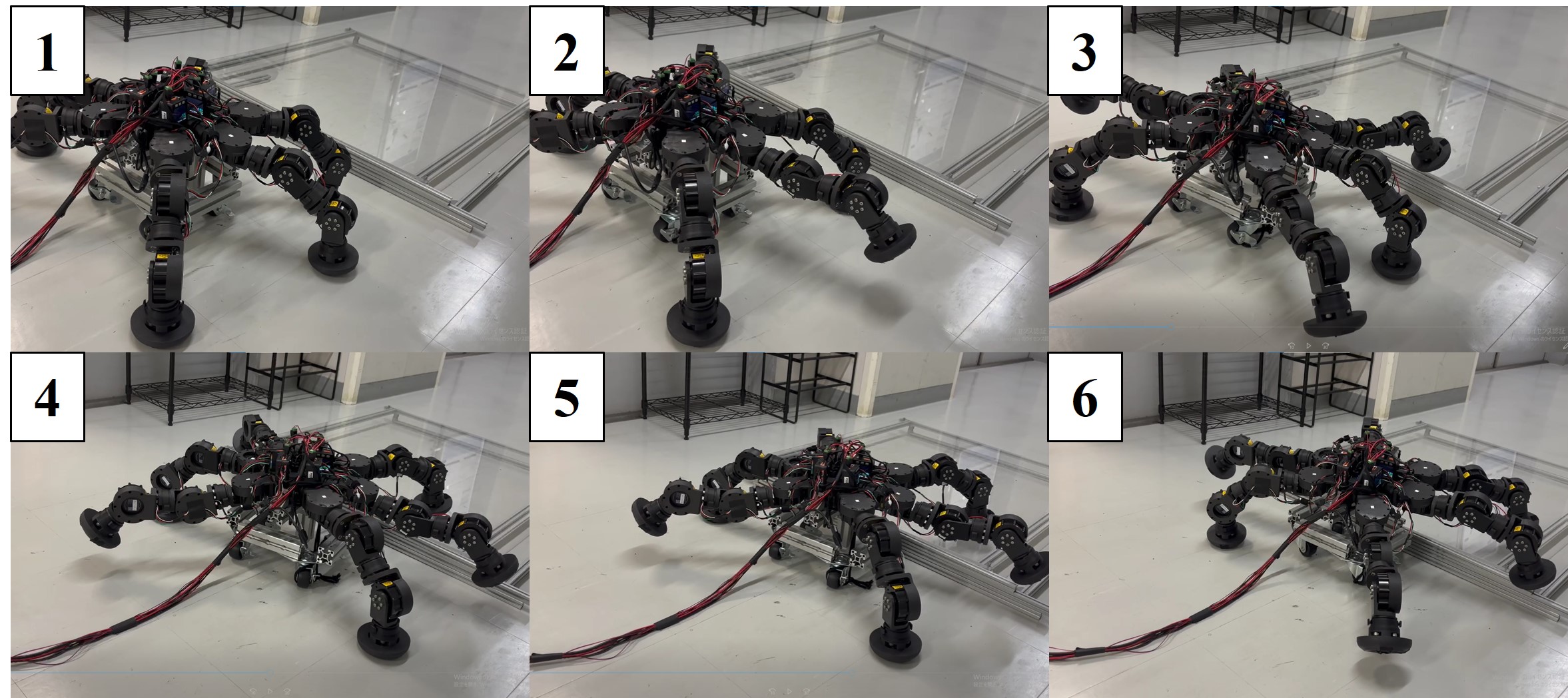}
    \subcaption{6-legged tripod (step, with external body support)}
  \end{minipage}
  \caption{Representative scenes of physical locomotion experiments. The red marker denotes a fixed reference point in the environment. External body support was used only in (d).}
  \figlab{5_exp_real_merged}
\end{figure}

\subsection{Feasibility Verification on Physical Robot}
Finally, we describe the results of applying the nondominated solutions obtained in the simulation to the physical robot. \tabref{5_exp_real_result_matrix} summarizes the success/failure results of the physical experiments. Each table entry represents the outcome of one physical trial; repeated trials were not conducted. \figref{5_exp_real_merged} illustrates representative scenes of the physical locomotion experiments. External support was used only in the physical step experiment shown in \figref{5_exp_real_merged}(d); the simulated step traversal in \figref{5_exp_uneven_gait_sequences}(b) was performed without external support.

On flat terrain, similar to the simulation, the 4-legged trot, 6-legged tetrapod, and 6-legged tripod gaits all succeeded in stable walking. Note that the 6-legged wave gait was not executed on the physical robot because it failed to achieve stable locomotion in the simulation. The simulated speeds reported below correspond to the specific candidates selected for transfer to the physical robots and therefore differ from the representative solutions discussed in Section~III-C. Regarding the locomotion speed, a tendency to be lower than the simulation values was observed. Specifically, for the 4-legged trot gait, the physical robot speed was $0.068\ \mathrm{m/s}$ compared to the simulated speed of $0.081\ \mathrm{m/s}$. Similarly, for the 6-legged tetrapod gait, the physical robot reached $0.068\ \mathrm{m/s}$, whereas the simulation result was $0.094\ \mathrm{m/s}$.

In the slope environment, the 4-legged trot gait ascended the slope without slipping, whereas the 6-legged robot slipped under the tested conditions. In the step environment, the 6-legged tripod gait failed to traverse the step without external support because body sinking reduced the effective swing-leg height. When the underside of the body was supported by a wheeled base, the foreleg and middle-leg stepping sequence was completed. These discrepancies suggest that the current simulation model does not fully account for the compliance of the 3D-printed body and the friction characteristics of the detachable joints under high load.

\section{Discussion}

\subsection{Optimization Trade-offs under Morphological Constraints}
The 10.5\% reduction in mean forward speed reflects the trade-off observed in the 6-legged tetrapod comparison: adding the joint-load objective reduced the mean maximum joint reaction force and increased the mean minimum stability. Separately, the 4-legged trot achieved a high minimum stability of 0.509 by selecting a large duty factor, which increased the duration of four-leg support.

The regression analysis showed a statistically significant positive association between swing height $H$ and the joint-load metric on flat terrain ($B=0.401$, $p=0.004$). A smaller $H$ reduces the vertical excursion of the swing legs and was associated with lower simulated reaction-force values at the modular interfaces. In contrast, the stride lengths and swing speeds of individual legs showed no statistically significant association, although larger values would be expected to increase inertial forces. A likely reason is that the swing speed was limited to at most $0.15\ \mathrm{m/s}$, making the motions nearly quasi-static, and that $f^{\text{load}}$ averages the reaction forces over the entire gait cycle, attenuating transient inertial peaks; thus, this result does not imply that these parameters have no effect on joint loads. In the slope and step conditions, no single decision variable showed a statistically significant association with the joint-load metric.

The observed dependence of locomotion feasibility on robot configuration, gait, and terrain is consistent with previous work on multi-legged modular robots with different configurations and levels of inter-module compliance~\cite{Mahkam2021}. This suggests that both the robot configuration and gait should be selected according to the terrain. Extending the optimization to morphological parameters, such as joint topology and link lengths, and evaluating a single configuration or terrain-aware gait switching on mixed terrains are promising directions for further improving terrain adaptability.

\begin{figure}[tb]
    \centering
    \includegraphics[width=\linewidth]{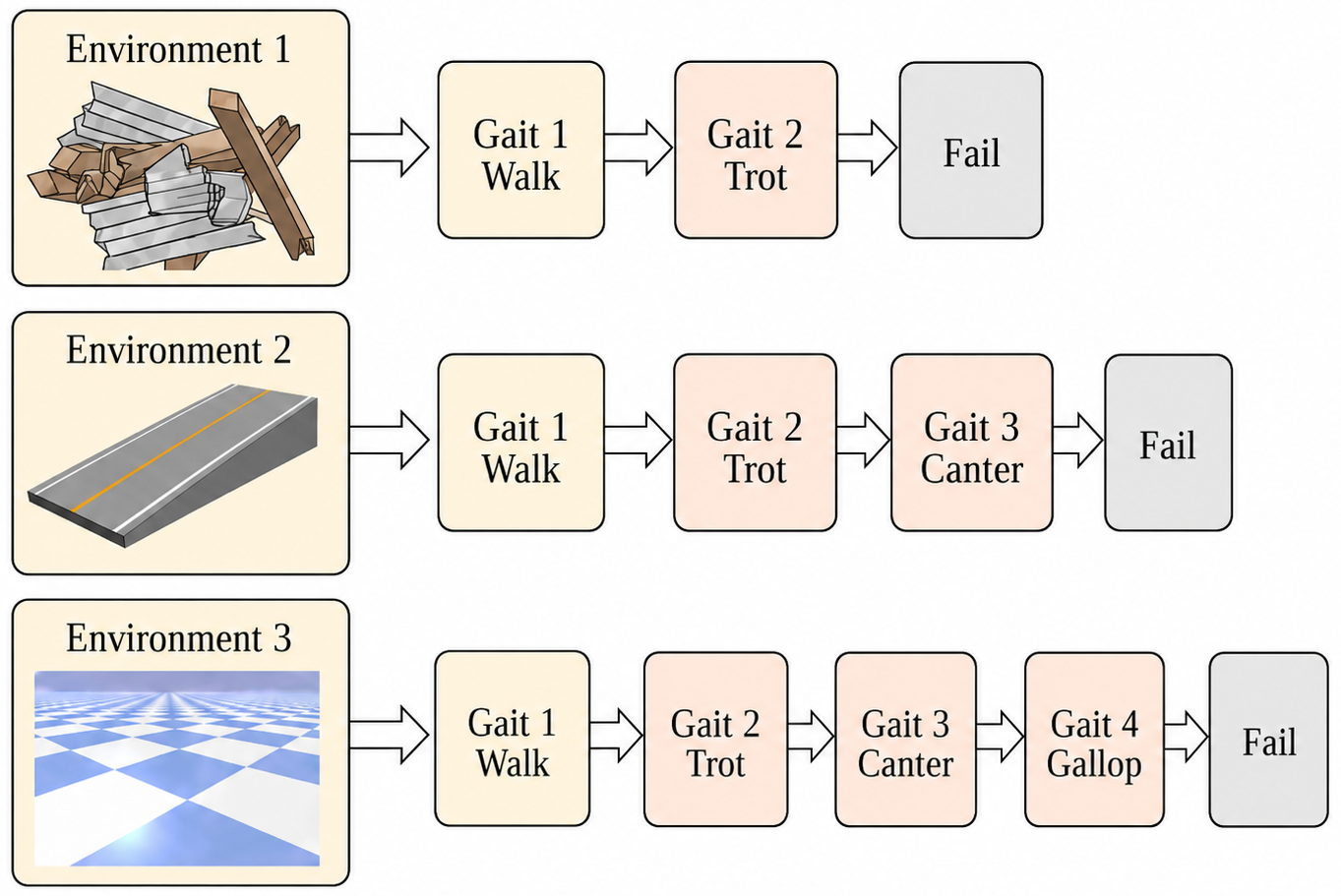}
    \caption{Conceptual illustration of a staged gait search with progressively increasing environmental difficulty for a four-legged modular robot.}
    \figlab{discussion_exploration_constraint}
\end{figure}

These observations motivate an environment-dependent staged search in which gait regions with more supporting legs are explored first and the search is gradually expanded toward gaits with more simultaneously swinging legs, as illustrated in \figref{discussion_exploration_constraint}. This idea is inspired by the relationship between support patterns and stability in mammalian symmetrical gaits~\cite{Cartmill2002}.

In a separate preliminary simulation using the 6-legged model in the slope environment, the staged search progressively relaxed the maximum number of simultaneously swinging legs from one to two and then three. The unconstrained search retained the same three objectives but imposed no limit on the number of simultaneously swinging legs, thereby searching the full gait space from the beginning. Each search was run for 10,000 gait evaluations. The unconstrained search did not discover a feasible gait within these evaluations, whereas the staged search obtained feasible nondominated solutions. This preliminary experiment differs from the evaluation in \tabref{5_exp_result_matrix}, in which the predefined wave, tetrapod, and tripod gait patterns were optimized separately. Therefore, the preliminary result indicates a difference in search efficiency and does not imply that no feasible 6-legged gait exists on the slope. Systematic evaluation across terrains and repeated runs remains future work.

\subsection{Sim-to-Real Gap and Hardware Limitations}
The lower locomotion speeds in the physical experiments coincided with visible body sinking, premature contact of the swing legs, and a smaller range of motion than in the simulation, in which the robot body and links were modeled as rigid bodies. These observations suggest that deformation of the 3D-printed body, joint backlash, and contact friction contributed to the sim-to-real gap; however, their individual contributions were not quantified.

The supported step experiment showed that the foreleg and middle-leg stepping sequence could be completed when body sinking was externally compensated. Unsupported step traversal was not achieved. Future work will measure body deformation and joint reaction forces on the physical robot and compare them with the simulation.

\section{Conclusion}
This study presented an NSGA-III-based gait optimization framework for balancing joint load and mobility in legged modular robots with detachable joints. In a flat-terrain comparison using the 6-legged tetrapod gait, adding the joint-load objective reduced the mean maximum joint reaction force and increased the mean minimum stability, while decreasing the mean forward speed. Simulations showed that locomotion feasibility depended on the combination of robot configuration, gait, and terrain. Physical experiments confirmed locomotion on flat terrain for the 4-legged trot and the 6-legged tetrapod and tripod gaits, and on the slope for the 4-legged trot. On flat terrain, swing height showed a statistically significant positive association with the joint-load metric. Future work will measure body deformation and joint reaction forces on the physical robot and refine the simulation model using these measurements to reduce differences between simulated and physical locomotion.

\section*{Acknowledgement}
This work was financially supported by the Nagamori Foundation.

\bibliographystyle{IEEEtran}
\bibliography{references}

\end{document}